\documentclass[letterpaper, 10 pt, conference]{ieeeconf}  

\IEEEoverridecommandlockouts                              

\usepackage{indentfirst}  
\usepackage{CJKutf8}
\usepackage[T1]{fontenc}
\usepackage{amssymb} 
\usepackage{cite}
\usepackage{placeins}
\usepackage{graphicx}
\usepackage{subcaption}
\usepackage{multicol}
\usepackage{tikz}
\usepackage{amsmath}
\usepackage{makecell} 
\usepackage{booktabs} 
\usepackage[normalem]{ulem}  
\usepackage{amsmath} 
\usepackage{amssymb}  
\usepackage{algorithm}
\usepackage{algorithmic} 
\usepackage[T1]{fontenc}
\usepackage{multicol}
\usepackage{multirow}
\usepackage{color}
\usepackage{cite}
\usepackage{placeins}
\usepackage{booktabs}
\usepackage{balance}
\makeatletter

\newcommand{\Rmnum}[1]{\expandafter\@slowromancap\romannumeral #1@}
\makeatother
\usepackage{epstopdf}
\AppendGraphicsExtensions{.tif}

\usepackage{hyperref}
\usepackage{cleveref}
\hypersetup{hidelinks}

\title{\LARGE \bf
LiDARDustX: A LiDAR Dataset for Dusty Unstructured Road Environments
}

\author{Chenfeng Wei$^{1*}$, Qi Wu$^{1*}$, Si Zuo$^{2}$, Jiahua Xu$^{1}$, Boyang Zhao$^{3}$, Zeyu Yang$^{2}$, Guotao Xie$^{2\dagger}$, Shenhong Wang$^{4\dagger}$ 
\thanks{$^{1}$Wuxi Intelligent Control Research Institute,HNU}
\thanks{$^{2}$Hunan University \quad{} $^{3}$Tsinghua University}
\thanks{$^{4}$Xi'an Jiaotong-Liverpool University}
\thanks{$^{*}$Equal contribution}
\thanks{$^{\dagger}$Corresponding author: \tt\small xieguotao90@hnu.edu.cn}
\thanks{$^{\dagger}$Corresponding author: \tt\small Shenhong.Wang@xjtlu.edu.cn}
}

\begin{document}
\maketitle

\begin{abstract}
Autonomous driving datasets are essential for validating the progress of intelligent vehicle 
algorithms, which include localization, perception, and prediction. However, existing datasets are predominantly focused on structured urban environments, which limits the exploration of unstructured and specialized scenarios, particularly those characterized by significant dust levels. This paper introduces the LiDARDustX dataset, which is specifically designed for perception tasks under high-dust conditions, such as those encountered in mining areas. The LiDARDustX dataset consists of 30,000 LiDAR frames captured by six different LiDAR sensors, each accompanied by 3D bounding box annotations and point cloud semantic segmentation. Notably, over 80\% of the dataset comprises dust-affected scenes. By utilizing this dataset, we have established a benchmark for evaluating the performance of state-of-the-art 3D detection and segmentation algorithms. Additionally, we have analyzed the impact of dust on perception accuracy and delved into the
causes of these effects. The data and further information can be accessed at: \textit{https://github.com/vincentweikey/LiDARDustX}.
\end{abstract}

\begin{figure}[!htbp]
    \centering
    \includegraphics[width=8.0cm]{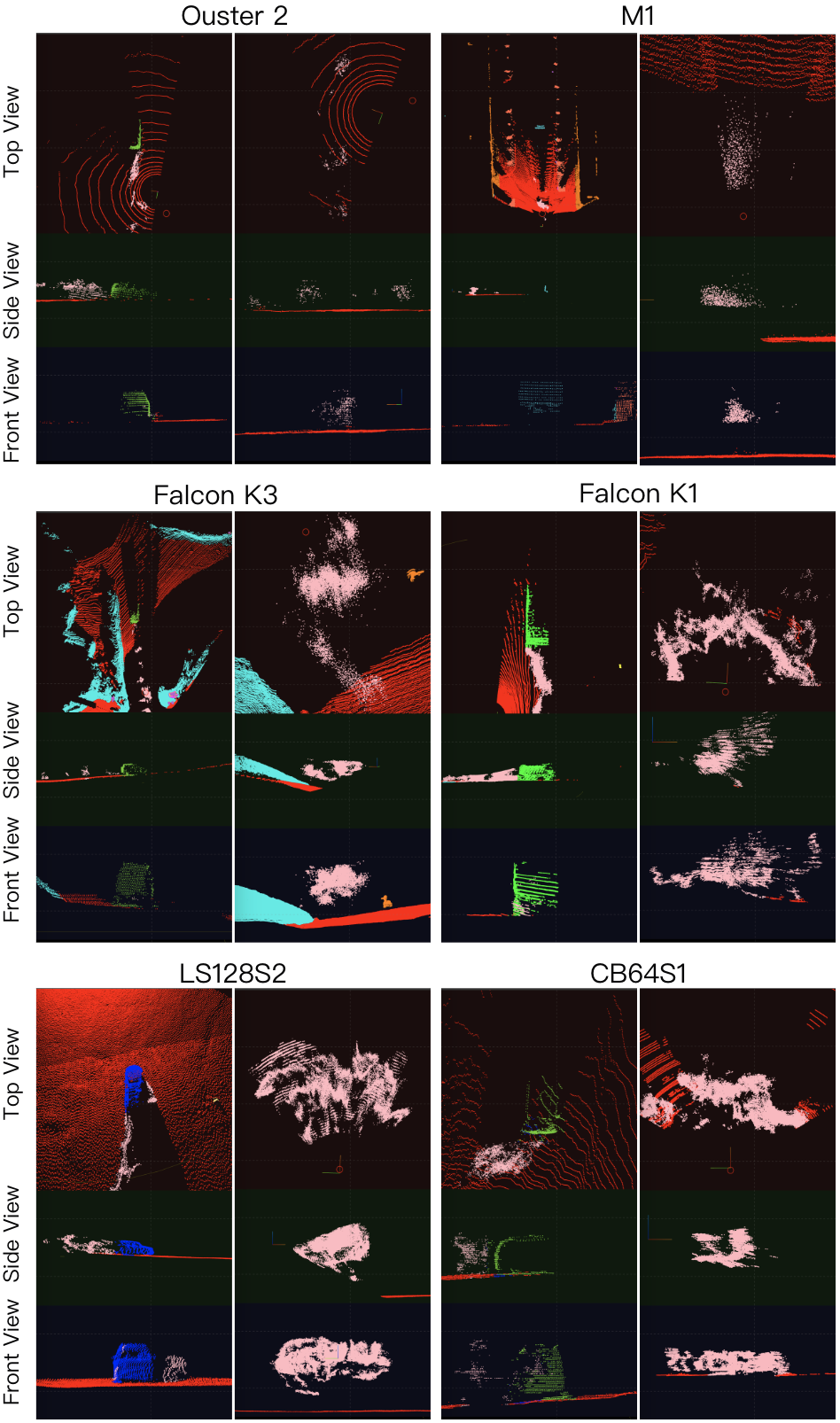}
    \caption{Overview of dust point clouds from 6 types of LiDAR in different perspectives, where pink represents dust, red represents the ground, light blue represents walls, and green/blue indicates different types of vehicles.
}
    \label{figure:1}
\end{figure}

\section{INTRODUCTION}
The evolution of autonomous driving technology has been significantly accelerated by datasets such as KITTI\cite{1}, nuScenes\cite{2}, Waymo\cite{3}, Argoverse\cite{4}, A*3D\cite{5}, A2D2\cite{6}, ONCE\cite{7}, and SemanticKITTI\cite{8}, which have provided extensive data for structured environments and a variety of tasks. However, there is a recognized deficiency in datasets that cater to the complexities of unstructured environments, particularly in mining areas where dust poses a significant challenge to perception tasks. The AutoMine\cite{9} dataset, while valuable for annotating mining scenarios, offers limited dust annotation. Consequently, LiDARDustX is introduced to bridge the existing gap. It features data from six diverse LiDAR sensors mounted on a truck, capturing a comprehensive range of dust scenarios. LiDARDustX is engineered to support the development of reliable autonomous systems capable of operating effectively in dusty conditions, thus expanding the capabilities of autonomous driving technology into more challenging and unstructured environments.

\begin{table*}[t]
    \centering
    \captionsetup{font=footnotesize, justification=raggedright} 
    \renewcommand{\arraystretch}{1.5} 
    \resizebox{\textwidth}{!}{
    \fontsize{10}{12}\selectfont 
    \begin{tabular}{cccccccccccc}
    \hline
             & Scenes & Frames & Classes & LiDAR Type & Object Detection & Object Tracking & 3D Boxes & Segmentation Label & Road & Dust & Attribute \\ \hline
    KITTI\cite{1} & 22 & 15k & 8 & 1 & $\checkmark$ & $\checkmark$ & 200k & $\times$ & Structured & $\times$  & Urban  \\ 
    nuScenes\cite{2} & 1k & 40k & \textbf{23} & 1 & $\checkmark$ & $\checkmark$ & 330k & $\checkmark$ & Structured & $\times$  & Urban  \\ 
    Waymo\cite{3} & \textbf{1.15k} & 230k & 4 & 2 & $\checkmark$ & $\checkmark$ & \textbf{12M} & $\checkmark$ & Structured & $\checkmark$  & Urban  \\ 
    Argoverse\cite{4} & 113 & 22k & 15 & 2 & $\checkmark$ & $\checkmark$ & 993k & $\times$ & Structured & $\times$  & Urban  \\ 
    A*3D\cite{5} & - & 39k & 7 & 1 & $\checkmark$ & $\checkmark$ & 230k & $\times$ & Structured & $\times$  & Urban  \\ 
    A2D2\cite{6} & - & 12k & 14 & 1 & $\checkmark$ & $\checkmark$ & 12k & $\checkmark$ & Structured & $\times$  & Urban/Highway  \\ 
    Once\cite{7} & - & \textbf{1M} & 5 & 1 & $\checkmark$ & $\checkmark$ & 417k & $\times$ & Structured & $\times$  & Urban  \\ 
    SemanticKITTI\cite{8} & 22 & 43k & 22 & 1 & $\times$ & $\times$ & - & $\checkmark$ & Structured & $\times$  & Urban  \\ 
    AutoMine\cite{9} & 70 & 18k & 9 & 2 & $\checkmark$ & $\times$ & 90k & $\times$ & Unstructured & $\checkmark$  & Mine  \\ 
    Ours & 180 & 30k & 16 & \textbf{6} & $\checkmark$ & $\checkmark$ & 300k & $\checkmark$ & Unstructured & \textbf{$\checkmark$} & \textbf{Mine/SQ} \\ \hline
    \end{tabular}
    }
    \caption{Comparison with existing representative 3D autonomous driving datasets. "-" means not mentioned, SQ: Sand Quarries. The term "Frames" refers specifically to "annotated frames".}
    \label{tab:Table1}
\end{table*}



LiDARDustX uniquely targets unstructured environments characterized by significant dust influence. This distinctive focus provides an essential resource for research in this domain, as shown in Fig. 1. The dataset's key strengths include:
\textbf{Diversity in LiDAR Sensors}: It features data from multiple LiDAR sensors, ensuring a range of data variability and representation.
\textbf{Comprehensive Dust Representation}: It captures a wide array of dust conditions from both engineering activities and natural events.
\textbf{High-Quality Annotations}: Professional annotators have meticulously labeled the data, including detailed 3D bounding boxes and point cloud semantic segmentation, guaranteeing accuracy and reliability.


Various deep learning methods were evaluated using the LiDARDustX dataset, focusing on single-task detection, segmentation, and multi-task models. Analysis revealed that the multi-task model excels in both detection and segmentation, especially in dusty conditions. It demonstrated improved detection capabilities and robustness against occlusions like vehicles and pedestrians.

The main contributions of this paper are as follows:
\begin{itemize}
\item 
A point cloud dataset for autonomous driving in unstructured road scenarios is released, specifically addressing the challenges posed by dusty environments.
\item 
A diverse array of LiDAR sensors is provided, with over 95\% of the frames exhibiting different forms of dust. Each data frame is accompanied by 3D bounding boxes and point cloud semantic segmentation annotations.
\item 
Experiments with various models on the dataset systematically analyzed the impact of dust from different sources on detection performance

\end{itemize}

\section{RELATED WORK}

The development of autonomous driving has been significantly advanced by early datasets KITTI\cite{1}, which introduced multi-sensor data for tasks such as object detection and semantic segmentation, setting a standard for the field. The nuScenes dataset\cite{2} further expanded the scope with its large-scale, 360-degree sensor data covering diverse scenarios and object types, offering a comprehensive benchmark for multi-task autonomous driving research.

The Waymo Open Dataset\cite{3}, released in 2019, offers high-resolution data from various environments, featuring multi-camera and LiDAR inputs alongside detailed annotations for object detection and tracking. It also includes sensor calibration data, aiding in sensor fusion and positioning studies. The Argoverse dataset\cite{4} provides 360-degree camera and LiDAR data with 3D annotations, focusing on dynamic object trajectory prediction for autonomous driving planning. Audi's A*3D\cite{5} and A2D2\cite{6} datasets, released in 2020, feature high-resolution multi-sensor data, with A2D2 emphasizing diverse scenarios and detailed annotations, suitable for evaluating multi-task autonomous driving algorithms.

The ONCE dataset\cite{7}, released in 2021, is an extensive collection with over 1 million annotated frames, encompassing urban, rural, and highway scenes. It offers detailed LiDAR point clouds, camera images, GPS/IMU data, and object annotations. The SemanticKITTI dataset\cite{8}, an extension of KITTI launched in 2022, specializes in semantic segmentation of LiDAR data, advancing research in this area. While most datasets focus on structured roads, the AutoMine dataset\cite{9} explores unstructured environments like open-pit mines, offering approximately 18K images and LiDAR data captured from various vehicles. Tab. \Rmnum{1} shows the comparison between the LiDARDustX dataset and other autonomous driving datasets.

\section{LIDARDUSTX DATASET}

In this section, the development of the dataset is discussed, with a focus on four key components: the data acquisition platform, LiDAR configuration, annotation techniques, and the statistical analysis of the data.

\subsection{Data Acquisition}

\begin{figure}[ht]
    \centering
    \includegraphics[width=8.4cm]{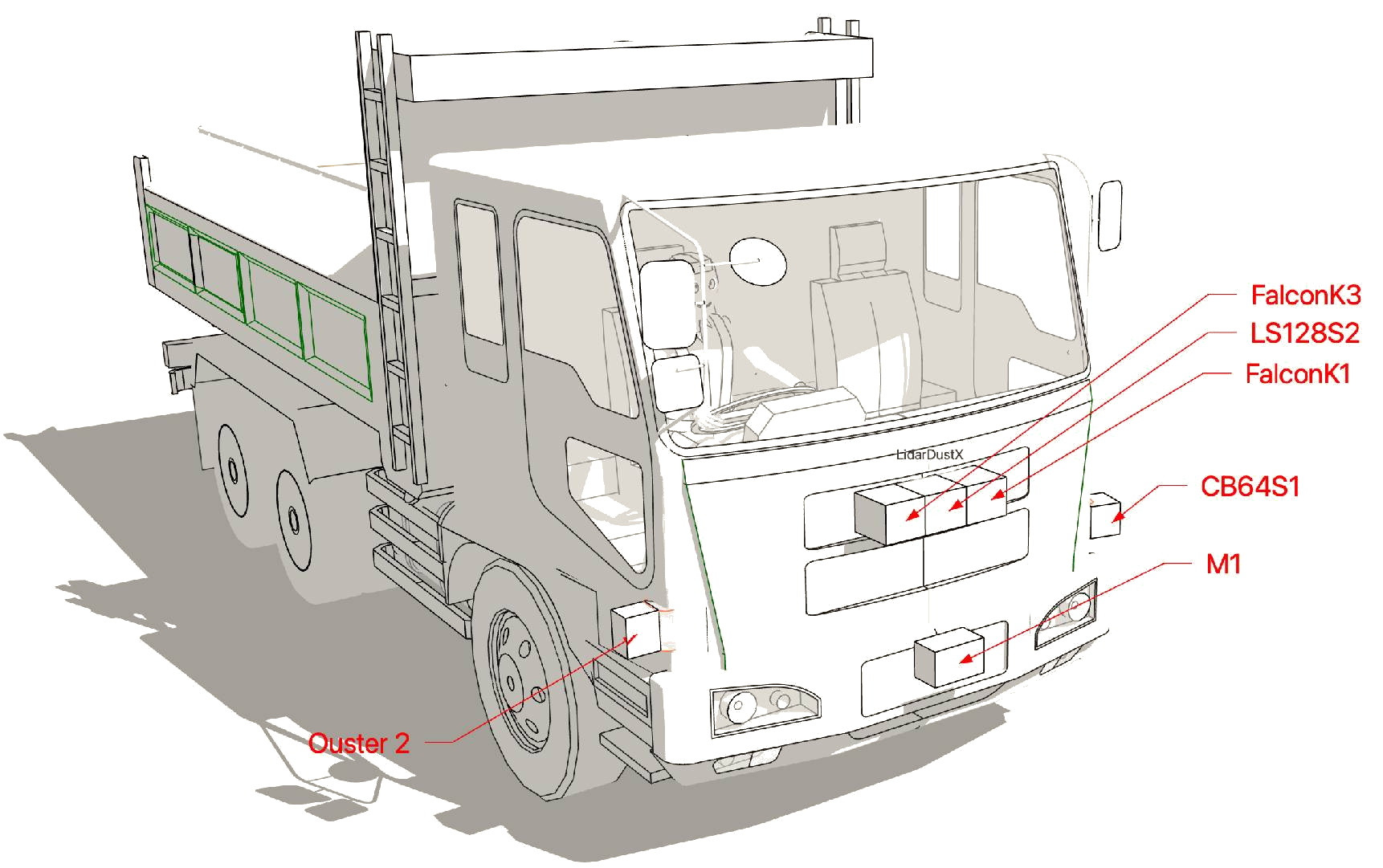}
    \caption{Collection Platform LiDAR Position Schematic.}
    \label{figure:2}
\end{figure}

\begin{figure*}[ht]
    \centering
    \includegraphics[width=\textwidth]{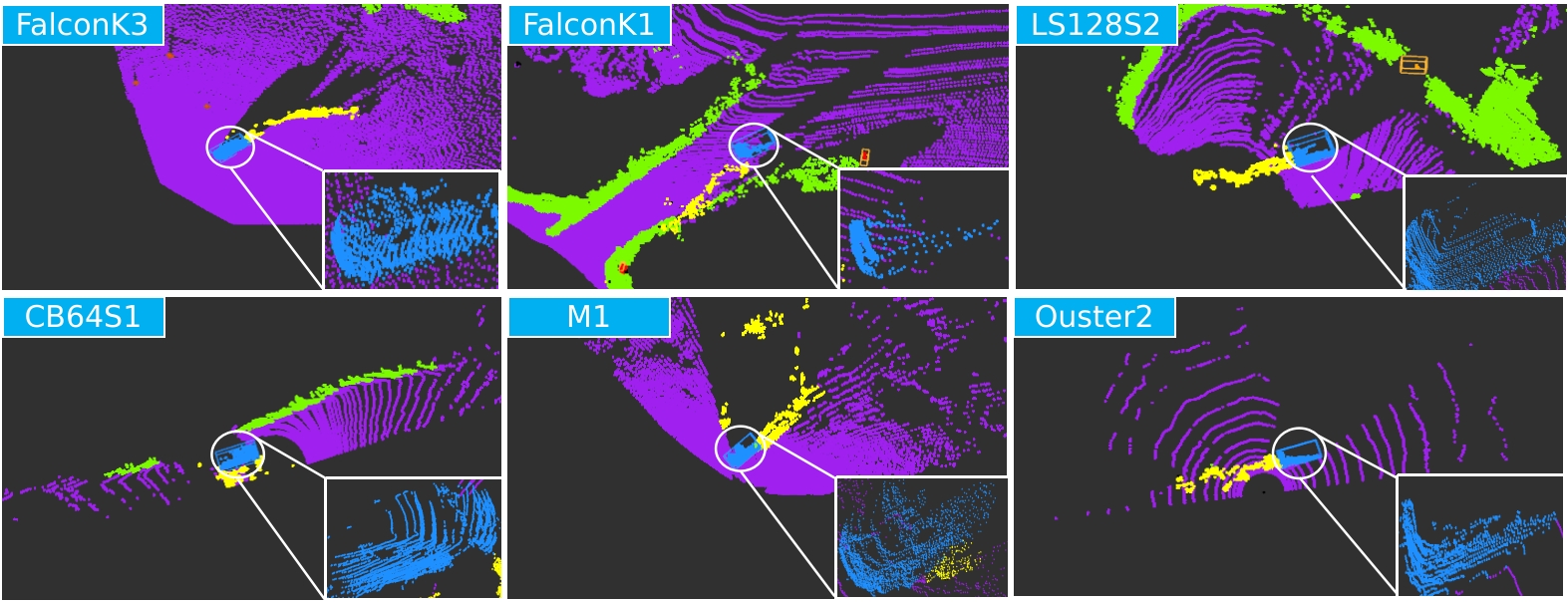}
    \caption{The LiDARDustX dataset showcases point clouds from various LiDAR sensors. The points in the cycle are zoomed in and shown in the white boxes for a better view.}
    \label{figure:31}
\end{figure*}

Collecting autonomous driving data in unstructured environments presents challenges due to fluctuating conditions, safety hazards, and legal considerations. The dataset was collected under clear daytime weather conditions, focusing on open-pit mines and sand quarries. Extensive data collection efforts were conducted across Inner Mongolia, Shanxi, and Gansu provinces in China. Characterized by severe natural conditions and ongoing engineering activities, these sites yielded genuine and valuable datasets that are crucial for evaluating the influence of dust on the perception systems of autonomous vehicles.

We used a TONGLI mining truck equipped with six different types of LiDAR sensors for data collection. From the raw sensor data, we carefully selected 200 dust-rich data sequences, each lasting approximately 20 seconds. These sequences encompass a variety of dust conditions, resulting from natural elements, blasting, excavation, and the movement of vehicles.

\subsection{Sensor Configuration}
The LiDARDustX dataset incorporates a variety of LiDAR sensors, including solid-state, hybrid solid-state, and mechanical types, with some sensors positioned at the same altitude while others are deployed at varying altitudes to capture diverse perspectives. Fig. 2 depicts the sensor configuration. The data collection platform is outfitted with a comprehensive sensor array, including Falconk3, LS128S2, FalconK1, and M1 devices positioned at the front, alongside ouster2 and CB64S1 equipment mounted on the side of the truck. Detailed specifications for each type of LiDAR are provided below:

1) \textbf{LS128S2} 120° horizontal, 25° (±12.5°) vertical viewing angle, a range of up to 180 meters and a measurement accuracy of ±2cm.

2) \textbf{CB64S1} 180° horizontal, -25° to 15° vertical viewing angle, a range of up to 100 meters and an accuracy of ±3 cm.

3) \textbf{FalconK1} 150-wire harness, a 120° horizontal, 25° vertical viewing angle, and a range of up to 250 meters.

4) \textbf{FalconK3} 300-wire harness, 120° horizontal, 20° vertical viewing angle and a maximum detection range of 300 meters.

5) \textbf{RS-LiDAR-M1} 120° horizontal, 25° vertical viewing angle, a range of 200 meters, and a measurement accuracy of ±5 cm.

6) \textbf{Ouster2} 64-wire harness, a vertical angle range of -7.9° to +7.9° and a range of 150 meters.


Operating at a 10Hz frequency, the LiDAR sensors in the LiDARDustX dataset provide high temporal resolution and precision in the collected data. Fig. 3 displays a curated set of point clouds from the dataset, illustrating the distinct shapes and structures identifiable by the various LiDAR sensors. For enhanced clarity of object details, the visualization focuses on cropped and magnified areas within each subfigure.

\subsection{Annotation Method}

A new annotation method has been introduced to simplify point cloud segmentation. It is designed to overcome the challenges associated with intersecting dust and irregular point clouds that complicate the use of 3D bounding boxes. Fig. 4 depicts this process.

1) \textbf{Point Cloud Stitching}: Calibrating six LiDAR sensors enables the stitching of their raw point clouds into a denser composite, making it easier for annotators to identify and improve the accuracy of point cloud feature calculations.

2) \textbf{Normal Vector Calculation}: Identify potential ground points by calculating the normal vectors of the point clouds. For each point in the point cloud, determine the normal vector by analyzing the surrounding local area and fitting a plane to proximate points. Normal vectors for ground points are predominantly oriented upwards, while those for non-ground points vary in direction.

3) \textbf{Region Growing Algorithm}: Through the utilization of potential ground points as seed points for the region growing algorithm, two distinct categories of point clouds are obtained: ground and non-ground.

4) \textbf{Clustering Non-Ground Points and Categorization}: The process involves clustering non-ground points, where labeled 3D bounding boxes serve to aggregate and categorize the data. Point cloud clusters that do not fall within the boundaries of any labeled 3D bounding box are considered as points to be ignored. This approach yields preliminary segmentation labels based on 3D bounding boxes alone.

5) \textbf{Segmentation Model Training}: The point cloud segmentation model, once trained with preliminary annotation data that contains rough labels, assigns a predicted label to each point, creating pseudo-labels that aid in the subsequent refinement process.

6) \textbf{Annotation Refinement}: Annotators refine the segmentation model's output by correcting misclassified segments with the aid of 3D bounding boxes, thereby attaining a high-precision point cloud segmentation annotation.

\begin{figure}[ht]
    \centering
    \includegraphics[width=8.4cm]{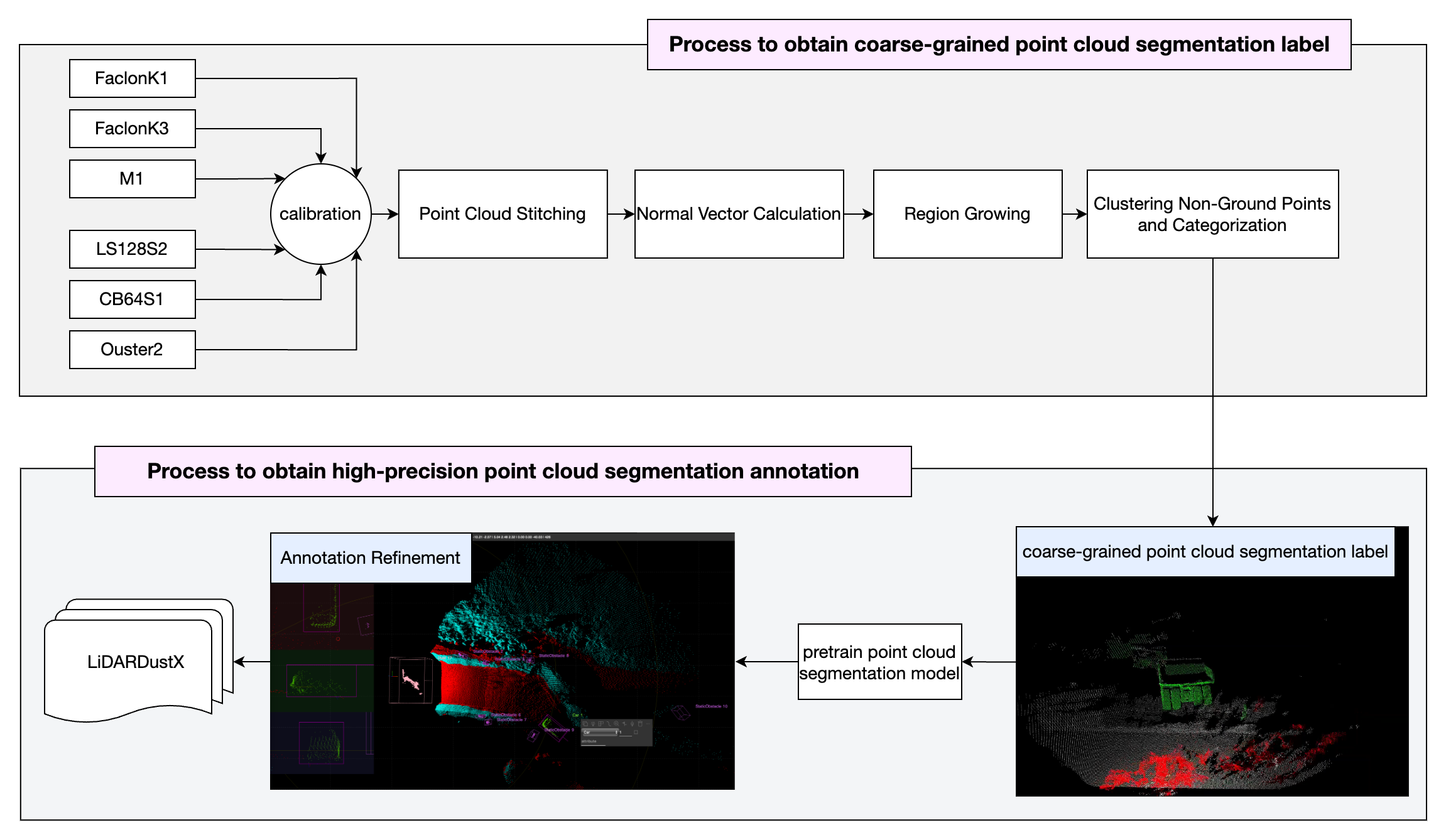}
    \caption{The methodologies for point cloud segmentation annotation.}
    \label{figure:32}
\end{figure} 

LiDARDustX dataset captures keyframes at 5 Hz, with each object annotated using a 7-DOF 3D bounding box specifying center coordinates (cx, cy, cz), dimensions (l, w, h), and the heading angle $\theta$. Each object is given a unique tracking ID. The dataset comprises 14 categories of 3D bounding boxes for objects like trucks and pedestrians, totaling 30,000 point cloud frames with over 300,000 annotated boxes. It also includes semantic labels for 16 classes and additional attributes like object ID for dynamic objects, essential for tasks in detection, tracking, and segmentation.

\subsection{Statistical Analysis}

\begin{figure}[ht]
    \centering
    \includegraphics[width=8cm]{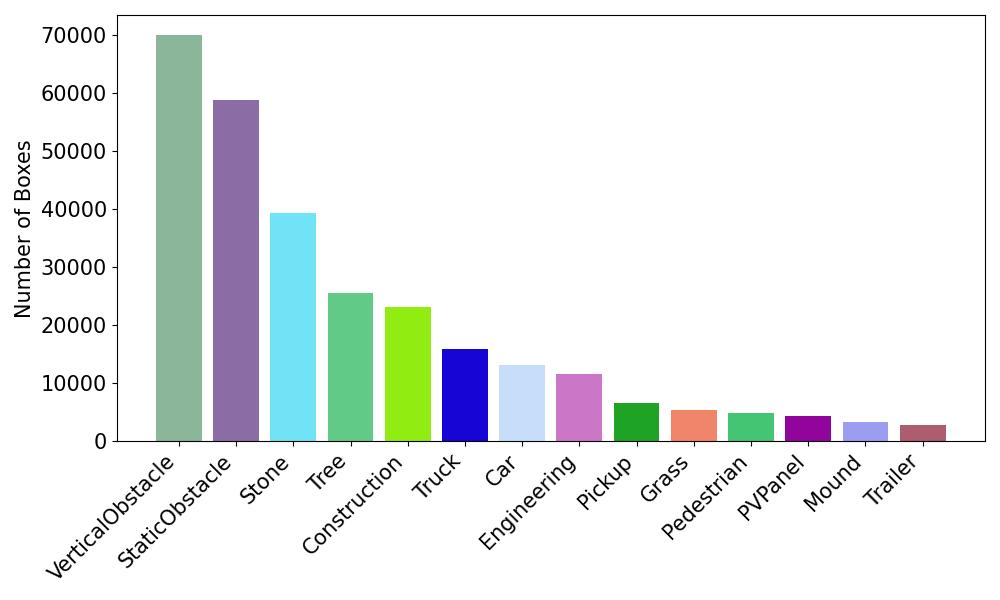}
    \caption{The number of 3D box annotations in different detection categories.}
    \label{figure:3}
\end{figure}

\begin{figure}[ht]
    \centering
    \includegraphics[width=9cm]{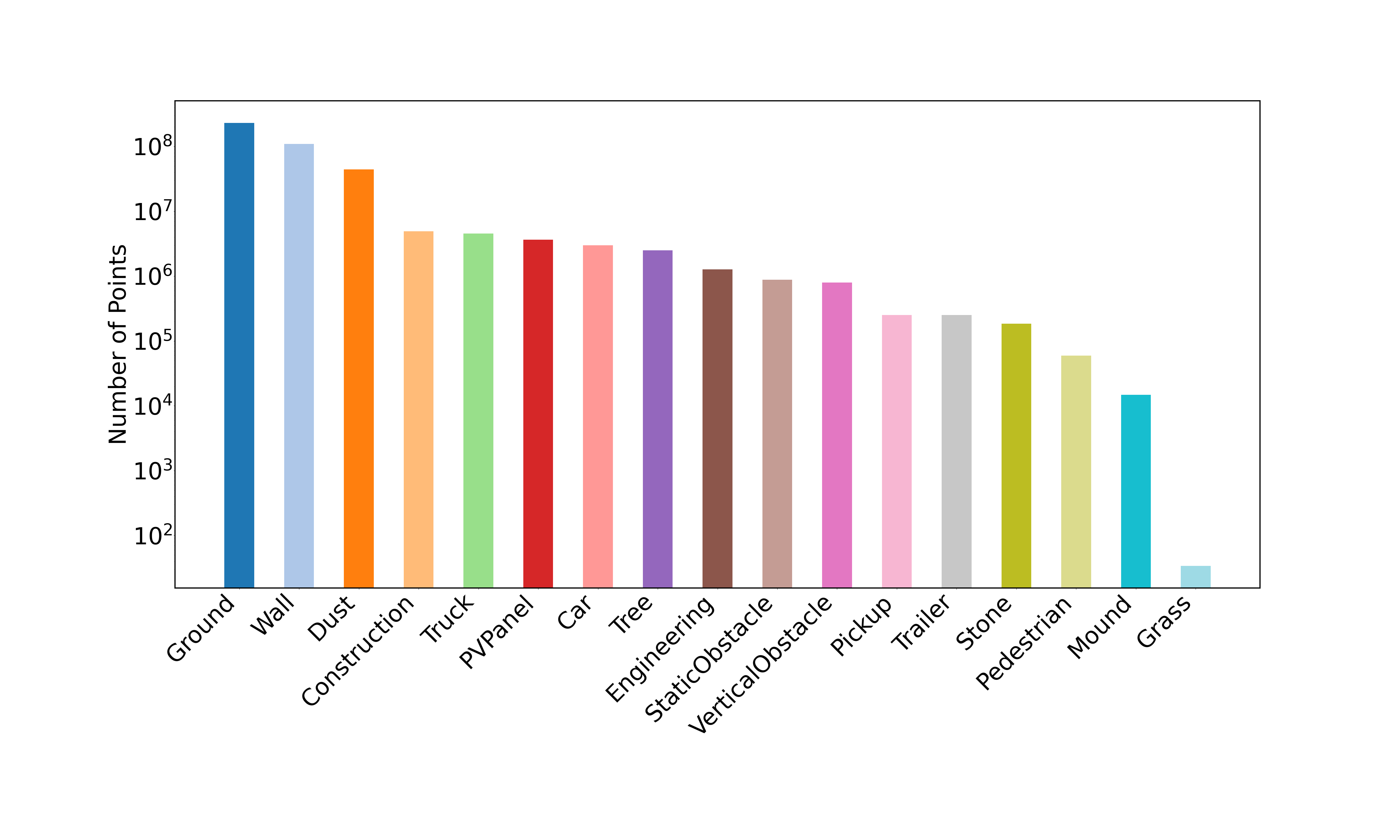}
    \caption{The number of points in different semantic categories.}
    \label{figure:4}
\end{figure}

\begin{figure}[ht]
    \centering
    \includegraphics[width=8cm]{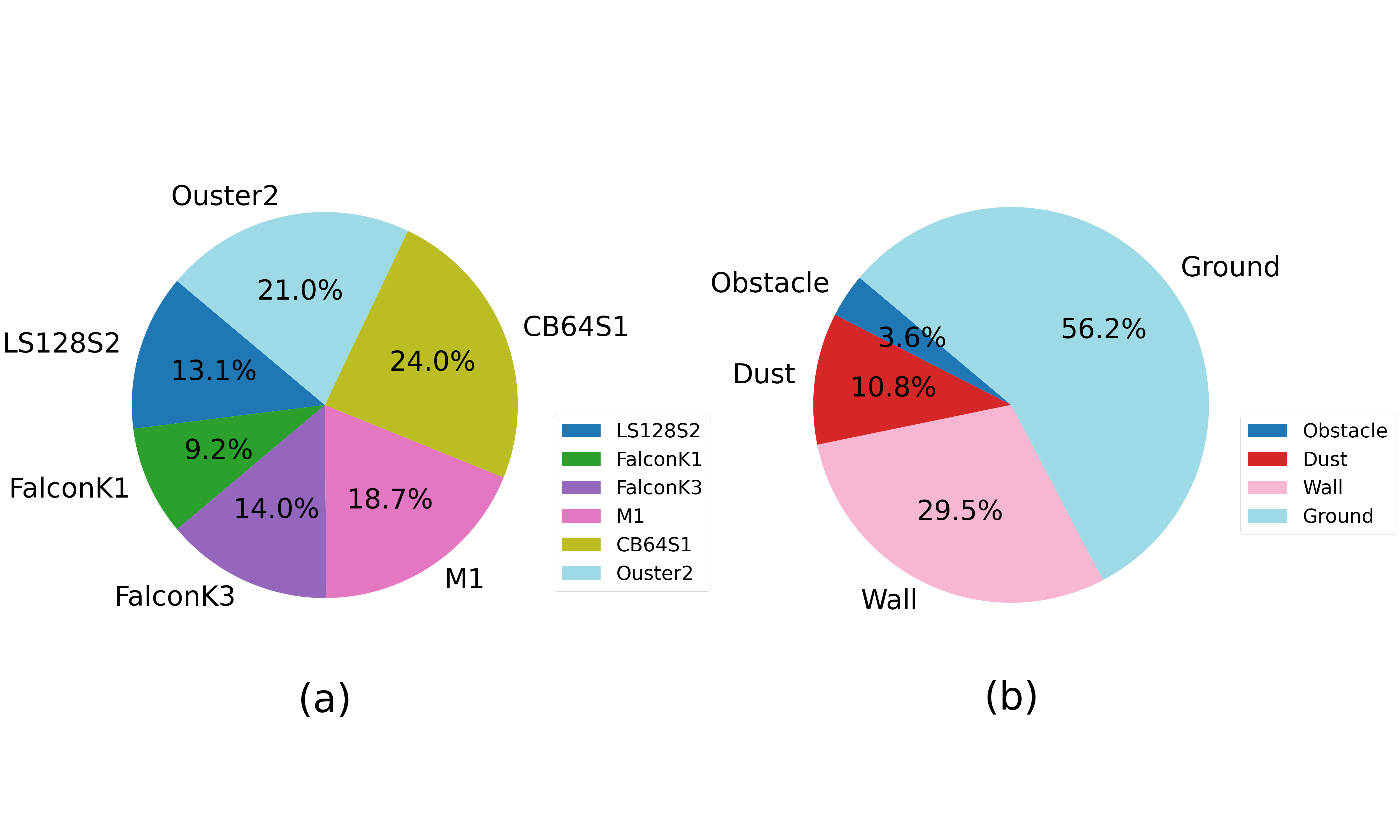}
    \caption{(a): The percentage of data frames under six different LiDAR sensor patterns. (b): The percentage of points belonging to various categories}
    \label{figure:5}
\end{figure}

In the 3D detection task, the dataset comprises 14 categories spanning various vehicles, pedestrians, and other objects, with statistical frequencies for each class presented (Fig. 5). The semantic segmentation schema encompasses a total of 16 classes, with Fig. 6 illustrating the annotated classes in LiDARDustX, including the point count for each. The ground, retaining walls, and dust categories are notably more prevalent in comparison to the others.

The fine-grained classes in LiDARDustX result in a severe class imbalance, marked by a 1:30 ratio between the least and most annotated classes, which motivates further community exploration of the long-tailed distribution. To facilitate point cloud segmentation analysis, categories like "Car," "Pedestrian," and "Building" have been merged into an "Obstacle" category. Additionally, the dataset's distribution across different LiDAR sensors is displayed, indicating a relatively even balance of LiDAR frames per sensor (Fig. 7(a)).


\begin{table*}[ht]
    \centering
    \captionsetup{font=footnotesize} 
    \renewcommand{\arraystretch}{1} 
    \resizebox{\textwidth}{!}{
    \begin{tabular}{ccccccccccc}
    \hline
              & mAP & Truck & Car & Pedestrian & mIOU & Ground & Dust & Obstacle & Wall \\ \hline
    PointPillars\cite{12}& 62.6 &  62.8  &  82.4 &  42.4 & - & - & - & -  \\ 
    PV-RCNN\cite{13}& 70.1 &  80.1 &  88.3 &  42.1 & - & - & - & - & - \\ 
    CenterPoint\cite{14} & 75.5 &  80.9 &  89.4 &  56.2 & - & - & - & - & - \\ 
    PillarNet \cite{15}& 77.4 &  80.5 &  89.5 &  62.1 & - & - & - & - & - \\ 
    VoxelNext\cite{16} & 78.3 &  81.4 &  90.1 &  63.5 & - & - & - & - & - \\ 
    TransFusion-L \cite{17}& 82.5 &  82.1 &  91.1 &  74.3 & - & - & - & - & - \\ \hline
    PointNet++ \cite{18} & - & - & - & - &  68.8 & 86.5 & 80.1 & 60.6 & 78.8  \\ 
    RandLA-Net\cite{20}  & - & - & - & - &  76.1 & 92.7 & 85.1 & 80.8 & 85.7  \\ 
    CeNet \cite{21}& - & - & - & - &  89.6 & 96.7 & 94.1 & 75.3 & 93.6   \\ 
    Cylinder3D\cite{19} & - & - & - & - &90.8 & 95.6 & 93.7 & 80.6 & 93.4 \\ \hline
    LisD  \cite{22}& \textbf{83.5} & \textbf{82.3} &  \textbf{91.5} &  \textbf{76.7} & \textbf{93.5} & \textbf{97.6} & \textbf{95.6} & \textbf{86.2} & \textbf{95.6} \\ \hline
    \end{tabular}
    }
    \captionsetup{font=normalsize} 
    \caption{Point cloud segmentation/detection model performance on the LiDARDustX dataset. The AP scores are measured at IOU = 0.7 thresholds for "TRUCK", and IOU=0.5 for "CAR" and "PEDESTRIAN" classes.}
    \label{tab:Table4}
\end{table*}

\subsection{Tasks \& Metrics}

LiDARDustX supports two major autonomous driving tasks: detection and segmentation. It includes a total of 180 sequences, which are split into a training set and a test set, with 70\% assigned to training and 30\% to testing. The test set is comprised of an equal amount of dusty and dust-free data, which can be used to assess the impact of dust on the model. The LiDARDustX detection task requires detecting 3 object classes(trucks, cars, and pedestrians) with 3D bounding boxes. These 3 classes are a subset of all 16 object classes annotated in LiDARDustX. We follow the evaluation metric in the NuScenes\cite{2} and SemanticKITTI\cite{8} benchmarks.

We use standard evaluation metrics for detection and segmentation, reporting mean Average Precision (mAP) and mean Intersection over Union (mIoU). For detection, mAP follows the official KITTI evaluation metric [1], calculated as the average precision (AP) over multiple recall positions, with IoU thresholds such as 0.7 for trucks and 0.5 for cars and pedestrians. For segmentation, mIoU\cite{10} evaluates the average overlap between predicted and ground truth regions across all classes, providing a comprehensive assessment of segmentation accuracy.


\section{BENCHMARK}

The generalization ability of several 3D object detection and segmentation baselines is evaluated on the LiDARDustX benchmark, with an exploration of the impact of dust on model performance. The detailed evaluation results are shown in Tab. \Rmnum{2}. 

\subsection{3D Object Detection Baselines}
Six widely used 3D detectors, reflecting mainstream methods, were implemented to thoroughly evaluate the performance of current 3D detection models on the LiDARDustX dataset. 

\textbf{\textbullet} \textbf{PointPillars} \cite{12} method converts point cloud data into vertical pillars, processes them with an MLP network, and utilizes the extracted features for detection via a 2D backbone network, streamlining point cloud processing.

\textbf{\textbullet} \textbf{PV-RCNN} \cite{13} integrates 3D voxel and point cloud data to learn features, employing 3D convolutions for global feature extraction and enhancing detection precision.

\textbf{\textbullet} \textbf{CenterPoint} \cite{14} is an anchorfree 3D-based detection framework that achieves high-accuracy target detection by directly predicting the position of the object's centroid.

\textbf{\textbullet} \textbf{PillarNet} \cite{15} enhances PointPillars with a deeper network and optimized feature extraction for improved accuracy and efficiency.

\textbf{\textbullet} \textbf{VoxelNext} \cite{16} enhanced VoxelNet \cite{23} variant, it boosts 3D detection speed and accuracy with efficient sparse convolution operations.

\textbf{\textbullet} \textbf{TransFusion-L} \cite{17} a LiDAR-only Transformer-based 3D detection framework.

\subsection{3D Semantic Segmentation Baselines}
For the evaluation of point cloud segmentation performance on the LiDARDustX dataset, several benchmark methods, including projection-based, voxel-based, and point-based schemes have been selected.

\textbf{\textbullet} \textbf{PointNet++} \cite{18} effectively improves point cloud processing by introducing a hierarchical feature learning mechanism.

\textbf{\textbullet} \textbf{Cylinder3D} \cite{19} significantly improves segmentation by projecting point cloud data to a cylindrical coordinate system and applying sparse convolutional networks to capture geometric features.

\textbf{\textbullet} \textbf{RandLA-Net} \cite{20} achieves fine segmentation by using random point cloud downsampling and local feature aggregation module to efficiently process large-scale point cloud data.

\textbf{\textbullet} \textbf{CENet} \cite{21} enhances the understanding of complex scenes by fusing multi-scale contextual information and global features and utilizing the self-attention mechanism.

\subsection{3D MultiTask Baseline}
\textbf{LiDAR MultiTask Baseline.}
In addition to single-task models, the performance of multi-task models was also evaluated on this dataset. The LiDAR multi-task model reduces computational demands by utilizing shared representation learning and enhances overall generalization by taking advantage of task synergy. 

\textbf{\textbullet} \textbf{LiSD} \cite{22} effectively integrates LiDAR semantic segmentation and object detection by leveraging a memory-efficient holistic information aggregation module and hierarchical structure.

\subsection{Dusty Challenge Results}
We combined dust-free and dusty point clouds for training and evaluation of multiple 3D object detection models. Performance was assessed on dust-affected and dust-free subsets, quantifying dust's impact on accuracy and revealing model robustness in visibility-impaired and dusty environments. The comprehensive results of these evaluations, including detailed performance metrics, are presented in Tab. \Rmnum{3}.

\begin{table}[h]
    \centering
    \captionsetup{font=footnotesize, justification=raggedright} 
    \begin{tabular*}{\columnwidth}{@{\extracolsep{\fill}} l c c c @{}}
    \toprule
        \textbf{Method} & \textbf{W/O Dust}
        & \textbf{W/Dust}  & \textbf{AP Drop} \\ \midrule
        PointPillars & 66.7 &  49.4  &  17.3 \\ 
        PV-RCNN & 71.5 &  56.3 &  15.2\\ 
        CenterPoint & 75.3 &  57.2 &  18.1\\ 
        PillarNet & 76.7 &  61.6 &  15.1 \\ 
        VoxelNext & 85.3 &  73.0 &  12.3 \\ 
        TransFusion-L & 85.6 &  75.7 &  9.9 \\ 
        \midrule
        LiSD & \textbf{86.5} & \textbf{76.4} &  \textbf{10.1} \\ 
        \bottomrule
    \end{tabular*}
    \caption{Performance Evaluation of Object Detection on LiDAR Data with and without Dust in the LiDARDustX Test Set (Average Precision at IOU = 0.7 for "TRUCK" and IOU = 0.5 for "CAR" and "PEDESTRIAN").}
    \label{tab:Table5}
\end{table}

\section{ANALYSIS}
Based on the above benchmarks, we have analyzed the critical characteristics and challenges that affect the performance metrics of our dataset.


\textbf{Robustness of Segmentation Tasks in Complex Environments.} In contrast to detection tasks, segmentation benchmarks show less variability in performance. Segmentation processes classify points individually, effectively managing local information. This approach enables the identification of points within the same category even when boundaries are obscured by dust, sustaining high accuracy. Notably, LiSD surpasses other segmentation methods in dusty conditions by refining segmentation features with detection features, markedly boosting segmentation performance.

\textbf{Collaborative Advantages of Multi-Task Models.} In dusty environments, multi-task models demonstrate superiority by leveraging features across various tasks, bolstering detection accuracy through segmentation supervision. This approach grants the model robustness in classifying points, especially under dusty conditions where single-task models might falter. The collaborative advantage of multi-tasking enhances overall performance, effectively eclipsing the capabilities of single-task models in the face of dusty disturbances.

\textbf{Impact of Dust on 3D Object Detection Performance.} The experiments distinctly reveal that dust substantially affects the accuracy of 3D object detection. In dust-free settings, models generally attain high AP, with LiSD and TransFusion-L excelling at 86.5\% and 85.6\% AP respectively. However, in dusty environments, a universal drop in detection accuracy is observed, notably a 17.3\% AP plummet for PointPillars, highlighting dust's detrimental impact. Conversely, LiSD and TransFusion-L, equipped with sophisticated feature processing, show greater stability in such conditions, retaining AP scores of 76.4\% and 75.7\%. 

\textbf{Detection Baseline Performance Variation.} A subjective examination was conducted on 100 randomly selected frames that exhibited detection errors to identify the underlying causes of these inaccuracies. It was notably found that frames containing both dust that occurred naturally and dust that was stirred up by vehicle movement were especially problematic. The detailed statistics are presented in Tab. \Rmnum{4}. Further analysis of these errors across various detection models revealed that natural dust primarily leads to an increase in false positives and negatives, without significantly affecting the orientation accuracy of the items detected. Conversely, dust induced by vehicles significantly degrades the precision of orientation and increases false positives, while false negatives are less frequently observed. This suggests that the effect of dust on detection models is dependent on its source, with each type of dust exerting a distinct influence on different detection metrics.

\begin{table}[h]
    \centering
    \captionsetup{font=footnotesize, justification=raggedright} 
    \begin{tabular*}{\columnwidth}{@{\extracolsep{\fill}} l c c c @{}}
    \toprule
        \textbf{Causes of dust formation} & \textbf{FP}
        & \textbf{FN}  & \textbf{Orientation error} \\ \midrule
        Vehicular activities & 21 &  25  &  0 \\ 
        Naturally occurring & 19 &  2 &  43\\ 
        \bottomrule
    \end{tabular*}
    \caption{Error Cause Statistics, FN: False Negatives, FP: False Positives)}
    \label{tab:Table55}
\end{table}




\section{Conclusion}
In this paper, the LiDARDustX dataset is introduced, encompassing tasks for detection and segmentation, baseline models, and corresponding results. The dataset, collected from a truck approved for testing on unstructured roads, comprises an extensive array of LiDAR dust data across different models. Employing our dataset as a benchmark provides a thorough analysis of the capabilities and constraints of contemporary 3D object detection and segmentation methodologies. Future aspirations include expanding the dataset with additional scenarios and integrating data from not only LiDAR but also camera and radar. The dataset has been made publicly accessible with the intention of fostering further advancements in 3D point cloud technology research. 
\bibliographystyle{ieeetr}
\bibliography{reference}

\end{document}